\documentclass{article}

\PassOptionsToPackage{numbers, compress}{natbib}

\usepackage{amsfonts}       
\usepackage{siunitx}        

    \usepackage[preprint]{neurips_2025}

\usepackage[utf8]{inputenc} 
\usepackage[T1]{fontenc}    
\usepackage{hyperref}       
\usepackage{url}            
\usepackage{booktabs}       
\usepackage{amsfonts}       
\usepackage{nicefrac}       
\usepackage{microtype}      
\usepackage{xcolor}         
\usepackage{graphicx}
\usepackage{amsmath}
\usepackage{longtable}

\title{The Birth of Knowledge: Emergent Features across Time, Space, and Scale in Large Language Models}

\author{%
  Shashata Sawmya$^{1}$ \quad Micah Adler$^{1}$ \quad Nir Shavit$^{1,2}$\\[4pt]
  $^{1}$Massachusetts Institute of Technology \\
  $^{2}$Red Hat, Inc.\\[4pt]
  \texttt{\{shashata,\,micah,\,shanir\}@mit.edu}
}

\begin{document}

\maketitle

\begin{abstract}
This paper studies the emergence of interpretable categorical features within large language models (LLMs), analyzing their behavior across training checkpoints (time), transformer layers (space), and varying model sizes (scale). Using sparse autoencoders for mechanistic interpretability, we identify when and where specific semantic concepts emerge within neural activations. Results indicate clear temporal and scale-specific thresholds for feature emergence across multiple domains. Notably, spatial analysis reveals unexpected semantic reactivation, with early-layer features re-emerging at later layers, challenging standard assumptions about representational dynamics in transformer models.
    
\end{abstract}

\section{Introduction}

Large Language Models (LLMs) and multimodal Vision-Language Models (VLMs) have become the standard computational tools across numerous applications, ranging from natural language understanding and generation to complex multimodal reasoning tasks. Their extensive deployment in both research and industry highlights their versatility and efficacy in handling a broad spectrum of computational problems. Despite this widespread usage, the internal mechanisms by which these models achieve their impressive performance remain largely opaque, resulting in their characterization as complex, "black-box" systems \cite{bommasani2021opportunities}.

Mechanistic interpretability has emerged as a promising research area aimed at dissecting the internal functioning of neural networks \cite{olah2020circuits}. By systematically identifying and describing the internal structures, this field seeks to uncover specific computational components—such as feature-level and circuit-level elements—that correspond to interpretable, human-understandable concepts. Among the methodological tools available for mechanistic interpretability, sparse autoencoders (SAEs) have proven particularly useful. SAEs impose explicit sparsity constraints on learned representations, thereby facilitating the extraction and interpretation of semantically meaningful features embedded within neural activations.

Previous studies have primarily quantified emergent behaviour by tracking aggregate metrics—task accuracy, generalization scores, or other benchmarks collected as models grow in size and training duration \cite{wei2022emergent, kaplan2020scaling, brown2020gpt3}. While informative, such evaluations reveal little about how the underlying computations themselves change over training or scaling .

\begin{figure}[h]
    \centering
    \includegraphics[width=\linewidth]{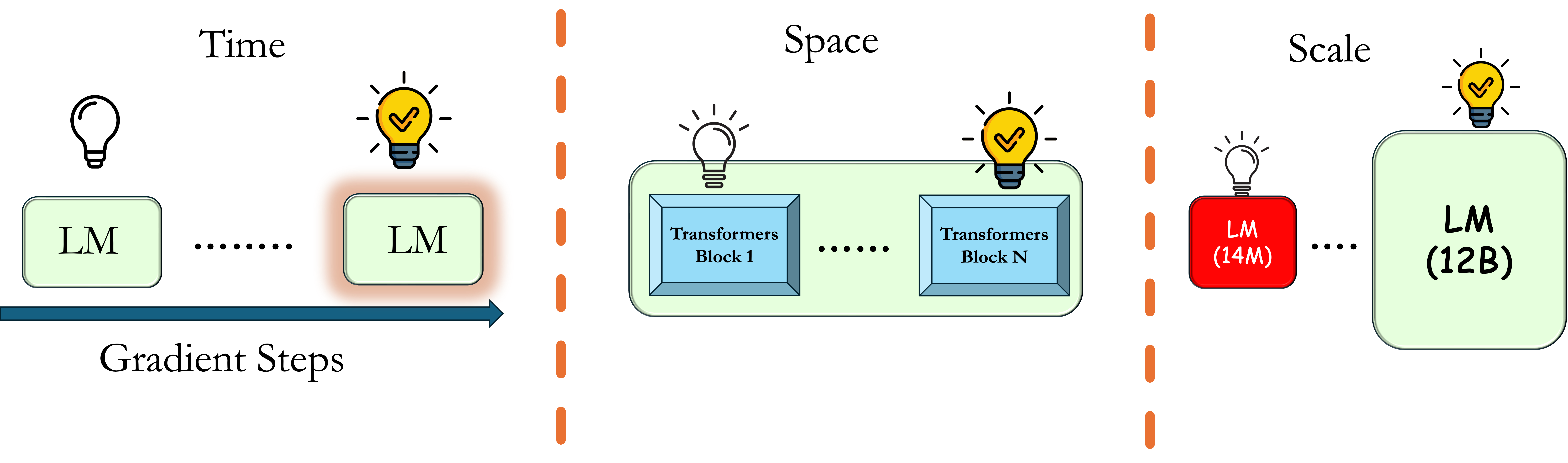}
    \caption{Axes of our emergent-knowledge probe. We track how interpretable, categorical features surface in a language model over time (training checkpoints), space (depth across transformer blocks), and scale (parameter count), progressing from sparse or absent concepts (grey bulbs) to rich representations (yellow bulbs)}
    \label{fig:}
\end{figure}

To address this gap we adopt an interpretability‑oriented methodology that probes the formation of semantically coherent features inside LLMs.  Our analysis targets three complementary axes: \emph{time} (checkpoints along the training trajectory), \emph{space} (positions in the transformer stack), and \emph{scale} (parameter count).  The prior hypothesis is that a model is comparatively ``un‑knowledgeable'' at its first gradient updates, within its earliest blocks, and when small in size; conversely, the density and specificity of category‑aligned features should increase as optimization proceeds, depth grows, and parameters multiply \cite{balagansky2025saematch, jawahar2019berthierarchy, jin2024conceptdepth, conmy2024autointerp, xu2024tracking}. Using sparse autoencoders we map when and where such features emerge, thereby charting the progressive structuring of internal representations in LLMs.

The primary contributions of this study are two-fold. First, we demonstrate the application of sparse autoencoders as mechanistic interpretability tools for uncovering emergent interpretable features within the residual streams of LLMs across \emph{three different axes}. Second, we conduct an extensive and fine-grained analysis of feature emergence and evolution in those dimensions across nine broad topical domains that span both the sciences and the arts. By probing into the feature dynamics associated with each topic, our analysis identifies the timing, location, and scale at which various interpretable features arise and mature within LLMs. 

The remainder of this paper is organized as follows. Section 2 and 3 presents the sparse autoencoder and autointerpretability methodology, experimental datasets, and model configurations. Section 4 examines the emergence of interpretable features along the training trajectory. Section 5 analyzes these features across the transformer stack, while Section 6 investigates their evolution under parameter scaling. Section 7 situates our work within the existing literature, and Section 8 offers concluding remarks.

\section{Background and Methods}
\label{sec:background}

\subsection{Model and Data}
\label{sec:model-data}

\textbf{Datasets.}  All experiments operate on the public \textsc{MMLU} test set (\num{14\,042} multiple-choice questions drawn from 57 academic subjects) \cite{hendrycks2021mmlu} and its harder extension \textsc{MMLU-Pro} (\num{12\,032} questions covering 14 broad categories) \cite{sun2023mmlupro}.  
Together the two benchmarks probe a model’s multitask general-knowledge competence over disciplines that span the sciences and the arts, e.g.\ Physics, Chemistry, Economics, Philosophy, History, and Ethics.  
For every item the question stem and all candidate answers are concatenated into a single string—omitting the correct letter—and passed through a language model once.  
The final-token hidden state is retained as the sample embedding used throughout this study.

\textbf{Pythia checkpoints for temporal, spatial, and scale analyses.}  
We adopt the \textsc{Pythia} suite of autoregressive transformers as a fully open substrate for mechanistic analysis \cite{biderman2023pythia}.  
For the \emph{temporal} investigation we track the 12-Billion-parameter model across 25 publicly released training checkpoints:
\[
\{0,1,2,4,8,16,256,512,1000,5000,10000,20000,\dots,140000,143000\},
\]
where the first ten steps give fine-grained coverage of early training and the remainder are spaced every 10 k updates up to near-convergence.  
For the \emph{scale} study we analyse all ten model sizes in the suite— 14M, 31M, 70M, 160M, 410M, 1B, 1.4B, 2.8B, 6.9B, and 12B parameters—each at its final checkpoint.  
For the \emph{spatial} study we focus on the 12 B model and extract embeddings from all 36 transformer blocks, allowing feature emergence to be mapped layer by layer.

\subsection{Sparse Autoencoders and AutoInterpretability}
\label{sec:sae}

\subsubsection{Architecture and Objective}
\label{subsec:sae-arch}

Let \(x \in \mathbb{R}^{d}\) be a residual–stream activation drawn from a transformer block.  
A sparse autoencoder maps \(x\) to an overcomplete latent space of width \(m>d\) in order to partition the original representation into finer-grained, potentially disentangled features.  
Formally, the encoder \(E \in \mathbb{R}^{m \times d}\) produces
\begin{equation}
    z = E\,x,
\end{equation}

after which a hard top-\(k\) operator, 
\(\tilde{z} = topk(z, k)\),
retains the \(k\) largest-magnitude coordinates and zeros the rest.  
The decoder \(D \in \mathbb{R}^{d \times m}\) reconstructs
\begin{equation}
    \hat{x}=D\,\tilde{z},
\qquad
\mathcal{L}_{\text{SAE}} = \|x - \hat{x}\|_2^2.
\end{equation}


We adopt OpenAI’s implementation of the \emph{top-\(k\) SAE} \cite{gao2024scaling}, which applies the masking step on every forward pass without an explicit sparsity penalty.  The strict cap of \(k\) active units accelerates dictionary formation: each optimisation step forces exactly \(k\) basis vectors to participate, so the overcomplete matrix \(E\) is populated with meaningful directions more quickly than in L\(_1\)- or KL-regularised variants, where sparsity emerges only through gradual weight adjustment.

In the standard top-\(k\) sparse auto-encoder only the \(k\) strongest latents update, so unused dictionary columns “die.’’  
The OpenAI implementation adds two small tricks: \textbf{multi-\(k\)}, which reruns the same forward pass with extra seeds of multi-\(k\) units receive gradient, and \textbf{aux-\(k\)}, where each silent neuron keeps a miss counter that, once past a threshold, promotes up to \(k_{\text{aux}}\) long-inactive units into the mask for one update before resetting.  
Both fixes periodically wake dormant features, cut dead-unit rates, and expand semantic coverage without changing the hidden width or inference cost.


\subsubsection{AutoInterp Pipeline}
\label{subsec:auto-interpret}

We adopt the external \textsc{AutoInterp} framework to generate and vet a natural-language label for every latent neuron in the SAE \cite{bricken2023towards}; our own contribution is limited to downstream use of the verified labels.

\vspace{0.3em}\noindent
\textbf{Label generation.}
For neuron \(j\) \textsc{AutoInterp} constructs two equal-sized example pools
\[
S^{+}_{j},\;S^{-}_{j}\subset\mathcal{D},
\qquad
|S^{+}_{j}|=|S^{-}_{j}| = n_{\mathrm{label}},
\]
where \(S^{+}_{j}\) contains data-points on which the neuron fires (post–top-\(k\) activation \(>0\)) and \(S^{-}_{j}\) contains the same on which it is silent.  
These text snippets are supplied to a teacher LLM with the instruction:  
\emph{“Describe the concept present in the first set but absent in the second.”}  
The LLM’s response is stored as the provisional label \(\ell_{j}\).

\vspace{0.3em}\noindent
\textbf{Verification via classifier metrics.}
A fresh, disjoint pair
\[
V^{+}_{j},\;V^{-}_{j},
\qquad
|V^{+}_{j}|=|V^{-}_{j}| = n_{\mathrm{verify}},
\]
is drawn using the same activation criterion.  
\textsc{AutoInterp} asks the teacher LLM whether \(\ell_{j}\) applies to each element of \(V^{+}_{j}\!\cup\!V^{-}_{j}\).  
Predictions are compared with ground truth (activation vs.\ no activation) to compute accuracy, precision, recall, and F1-score.  
Neuron \(j\) is deemed \emph{interpretable} if its F1-score exceeds a preset threshold \(\tau_{\mathrm{F1}}\); otherwise it is excluded from further analysis.

This automatic label–verify loop scales linearly with the number of neurons while requiring only two small sample budgets, \(n_{\mathrm{label}}\) for hypothesis formation and \(n_{\mathrm{verify}}\) for metric-based validation.

\subsection{Design Choices}
\label{sec:design-choices}

\paragraph{Modified AutoInterp.}
We follow the \textsc{AutoInterp} framework (Section~\ref{subsec:auto-interpret}) with two pragmatic deviations.  
First, the \emph{label} prompt is constructed from the top-activation set only; we use \(n_{\mathrm{label}}=10\) examples and omit the non-activating counterparts.  
Second, the \emph{verification} prompt retains the original balanced split, drawing \(n_{\mathrm{verify}}=5\) activating and \(5\) non-activating samples to compute classifier metrics (accuracy, precision, recall, F1) for each candidate label.

\begin{figure}[h]
    \centering
    \includegraphics[width=0.9\linewidth]{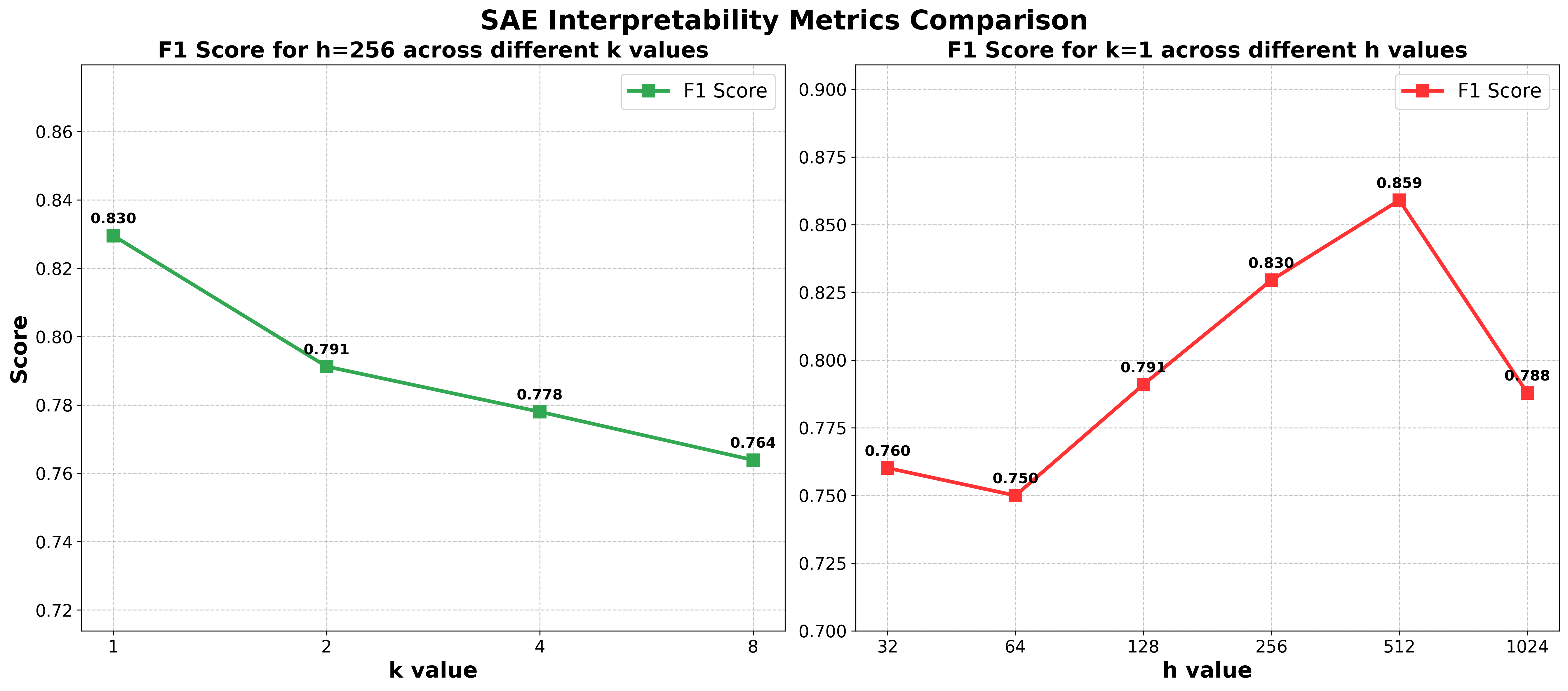}
    \caption{Hyperparameter sweep for sparse–autoencoder interpretability.  
Left: mean F1‐score as the activation budget \(k\) varies with width fixed at \(h=256\); right: mean F1‐score as the latent width \(h\) varies with \(k=1\). 
The optimal setting for our data is \(k=1,\,h=512\), which maximises mean F1.}
    \label{fig:k_h_sweep}
\end{figure}

\paragraph{Selecting the activation budget \(k\).}
With the hidden dimension fixed at \(h=256\) we tested \(k\in\{1,2,4,8\}\).  
The mean F1-score across all labelled neurons decreased monotonically with larger \(k\), and \(k=1\) achieved the highest score (Figure~\ref{fig:k_h_sweep}, left).  
We therefore set \(k=1\) for the remainder of the study.

\paragraph{Selecting the latent width \(h\).}
Holding \(k=1\) constant, we varied the latent dimensionality
\[
h\in\{32,64,128,256,512,1024\}.
\]
The mean F1-score peaked at \(h=512\) (Figure~\ref{fig:k_h_sweep}, right).  
Dead-latent incidence was non-negligible; the number of neurons that ever activated during training was
\(\{32,54,120,170,211,122\}\) for the six \(h\) values, respectively.  
Balancing interpretability (via F1) against parameter count, we chose \(h=512\).

\paragraph{Why \(h \ll d\).}
The residual-stream dimensionality of the Pythia-12B model is \(d=5120\).  
We intentionally select \(h\ll d\) for two reasons.  
First, the combined benchmark contains only \(57+14\) nominal subject labels, several of which overlap (e.g.\ \textit{high\_school\_physics} vs.\ \textit{college\_physics}); the effective concept inventory is therefore far smaller than \(d\).  
Second, this study focuses on the \emph{emergence} of coarse interpretable features rather than full disentanglement of polysemantic directions, making an overcomplete basis of size \(h=512\) sufficient for our analysis goals.

\section{EyeSee: A Framework for Probing Categorical Concepts}
\label{sec:eyesee}

Modern benchmarks such as \textsc{MMLU} and \textsc{MMLU‐Pro} are organised around textbook disciplines (e.g.\ \textit{Physics}, \textit{History}).  
If an LLM truly internalises these domains, one would expect dedicated latent directions to emerge that activate whenever the input concerns a given subject.  
Our goal is therefore to \emph{distil} the full set of SAE neurons down to those that reliably represent such high-level categories and to track their behaviour across time, space, and scale.

\paragraph{High-fidelity neuron pool.}
Running \textsc{AutoInterp} on the selected SAE (\(h\!=\!512,\,k\!=\!1\)) yields a verification F1-score for every latent label.  
We keep only neurons whose score exceeds \(0.9\), forming a trusted set \(\mathcal{N}_{\!\mathrm{hi}}\).

\begin{table}[ht]
\centering
\resizebox{\textwidth}{!}{%
\begin{tabular}{lcccc}
\toprule
\textbf{Subject} & \textbf{Neuron ID} & \textbf{AutoInterp Label} & \textbf{Cos.\ Sim.} & \textbf{F1} \\
\midrule
Biology      & 38  & Natural Phenomena and Processes      & 0.377 & 0.91 \\
             & 295 & Genetic Variation Source             & 0.366 & 1.00 \\
             & 269 & Human-related Processes              & 0.347 & 0.91 \\
\hline
Chemistry    & 194 & Paper-related Chemistry              & 0.491 & 1.00 \\
             & 125 & Iron and Silver Chemistry            & 0.406 & 1.00 \\
             & 38  & Natural Phenomena and Processes      & 0.385 & 0.91 \\
\hline
Physics      & 78  & Newton’s Laws Applications           & 0.513 & 1.00 \\
             & 195 & Mathematical Problem Solving         & 0.425 & 1.00 \\
             & 12  & Real-world Mathematical Applications & 0.401 & 1.00 \\
\hline
Mathematics  & 195 & Mathematical Problem Solving         & 0.707 & 1.00 \\
             & 400 & Logical and Mathematical Concepts    & 0.590 & 0.91 \\
             & 12  & Real-world Mathematical Applications & 0.549 & 1.00 \\
\hline
Economics    & 367 & Business and Economic Dynamics       & 0.600 & 1.00 \\
             & 55  & Human Behavior and Decision-Making   & 0.485 & 0.91 \\
             & 218 & Economic Growth Factors              & 0.480 & 1.00 \\
\hline
History      & 4   & Social Dynamics and Influence        & 0.387 & 1.00 \\
             & 58  & Comparative Analysis                & 0.322 & 0.91 \\
             & 446 & Ethical and Cultural Positions       & 0.320 & 1.00 \\
\hline
Law          & 510 & Legal Decision-Making Criteria       & 0.454 & 1.00 \\
             & 57  & Warranty Types in Law                & 0.377 & 1.00 \\
             & 305 & Decision-Making in Institutions      & 0.345 & 1.00 \\
\hline
Philosophy   & 350 & Ethical and Philosophical Concepts   & 0.606 & 0.91 \\
             & 285 & Socratic Philosophy Concepts         & 0.477 & 1.00 \\
             & 446 & Ethical and Cultural Positions       & 0.462 & 1.00 \\
\hline
Business     & 367 & Business and Economic Dynamics       & 0.548 & 1.00 \\
             & 456 & Distribution Channels                & 0.419 & 1.00 \\
             & 55  & Human Behavior and Decision-Making   & 0.314 & 0.91 \\
\bottomrule
\end{tabular}}
\vspace{0.6em}
\caption{Top three high-fidelity concept neurons per subject, selected by cosine similarity $\ge 0.3$ between the subject name and the AutoInterp label in MPNet embedding space.  The additional \textit{F1} column reports the verification fidelity for each neuron.  Complete ranked lists are provided in Appendix~A.1.}
\label{tab:concept-neurons}
\end{table}

\paragraph{Query-driven concept matching.}
We begin with the free-text neuron labels associated with the trusted high-fidelity pool \(\mathcal{N}_{\mathrm{hi}}\) and encode each label once with the sentence-embedding model \texttt{all-mpnet-base-v2}, thereby building a lightweight vector database.  
Whenever a query subject (e.g.\ \textit{Physics}, \textit{History}) is posed, we encode that query with the same model and retrieve the neuron-label vectors that lie closest to it in the embedding space.  Cosine similarity acts as the nearest-neighbour criterion that links query subjects to the neurons whose labels express the most semantically aligned concepts.  The precise formulation is given below.

Denote by $\varphi$ the \texttt{all-mpnet-base-v2} embedding function.  
For each neuron \(j\!\in\!\mathcal{N}_{\!\mathrm{hi}}\) we compute \(v_{j}=\varphi(\ell_{j})\).  
Given a subject \(q\in\mathcal{Q}\) with embedding \(u_{q}=\varphi(q)\), the cosine similarity is
\begin{equation}
s_{j,q} \;=\;
\frac{u_q^{\top} v_j}{\lVert u_q \rVert_2 \,\lVert v_j \rVert_2}.
\end{equation}

Fixing a threshold \(\tau=0.3\), the candidate concept set for \(q\) is
\[
\mathcal{N}_{q}=\{\,j\in\mathcal{N}_{\!\mathrm{hi}} : s_{j,q}\ge\tau\,\}.
\]
Neurons in \(\mathcal{N}_{q}\) are ranked by \(s_{j,q}\).  
Table~\ref{tab:concept-neurons} reports the three highest-similarity neurons per subject for illustration; complete ranked lists appear in Appendix~A.1.  
These subject-aligned neurons serve as probes when analysing feature emergence along the temporal, spatial, and scaling dimensions in subsequent sections.

\section{Temporal Emergence of Categorical Knowledge in \textsc{Pythia-12B}}
\label{sec:time-concepts}

\begin{figure}[h]
    \centering
    \includegraphics[width=\linewidth]{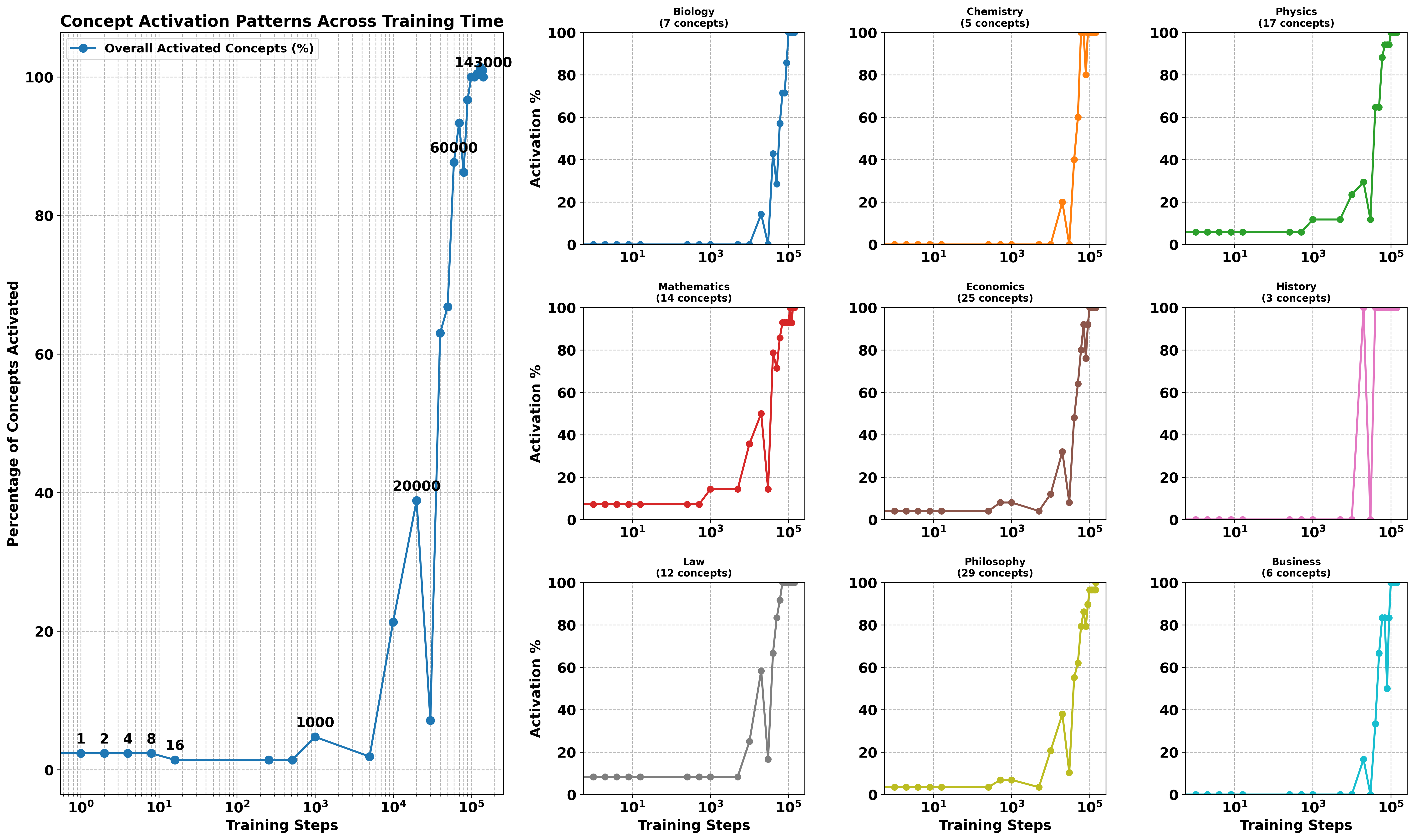}
    \caption{Activation patterns of categorical concepts in a 12B-parameter language model across training checkpoints. The left panel illustrates the global activation trajectory, while panels on the right display domain-specific emergence patterns, highlighting distinct activation timings for various knowledge concepts.}
    \label{fig:concept_time}
\end{figure}

Using the SAE trained at the last training checkpoint, we ask the following question, \textit{What concepts in the form of feature neurons are activated by LM embeddings curated from other gradient steps}? Figure \ref{fig:concept_time} tracks the percentage of concepts that become active as the model traverses its training trajectory (horizontal axis is \textit{log}-scaled steps). We begin with the global curve (far-left panel) and then drill down into ten representative knowledge areas (right grid).

\paragraph{Global pattern.}
For the first \(10^{3}\) optimization steps fewer than \(3\%\) of concepts are active.  
A first increase of \(+19.4\) percentage points (pp) appears at \(5\,000\) steps.  
Two larger increments follow:
\begin{enumerate}
    \item \(10\,000\!\rightarrow\!20\,000\) steps: \(+17.5\) pp,
    \item \(30\,000\!\rightarrow\!40\,000\) steps: \(+55.9\) pp, following a \(-31.8\) pp dip in the \(20\,000\!\rightarrow\!30\,000\) interval.  
          This dip may reflect either (i) a re-organisation of feature representations or  
          (ii) a temporary reduction in gradient-driven optimization efficacy before training resumes.
\end{enumerate}
After \(40\,000\) steps the curve continues to rise and exceeds \(99\%\) by the final checkpoint at \(143\,000\) steps.

\paragraph{Domain-specific activation patterns.}
Inspection of Figure~\ref{fig:concept_time} reveals two broad temporal patterns.

\begin{itemize}
\item \textbf{Early-onset domains} (Physics, Mathematics, Economics, Law, Philosophy).  
      These subjects register non-zero activations from the very first optimisation steps and rise gradually, then surge after \(\sim\!3\times10^{4}\) steps to exceed \(60\!-\!90\%\).  Their early presence hints at the high frequency of numerical, symbolic, and formal language in the training text corpus.
\item \textbf{Late-onset domains} (History, Biology, Chemistry, Business).  
      Activations remain at \(0\%\) until roughly \(10^{4}\) steps, after which they climb steeply—often in a single burst—to reach near-saturation between \(3\times10^{4}\) and \(6\times10^{4}\) steps.  These topics appear to depend on higher-level contextual structures that only stabilise once lower-level patterns have been learned.
\end{itemize}

 By the final checkpoint, all nine subjects approach full activation, indicating a progressive—not instantaneous—accumulation of domain knowledge during pre-training.



\section{Analysis of Representation Space Across Model Layers}
\label{sec:space_ana}
\begin{figure}[h]
    \centering
    \includegraphics[width=\linewidth]{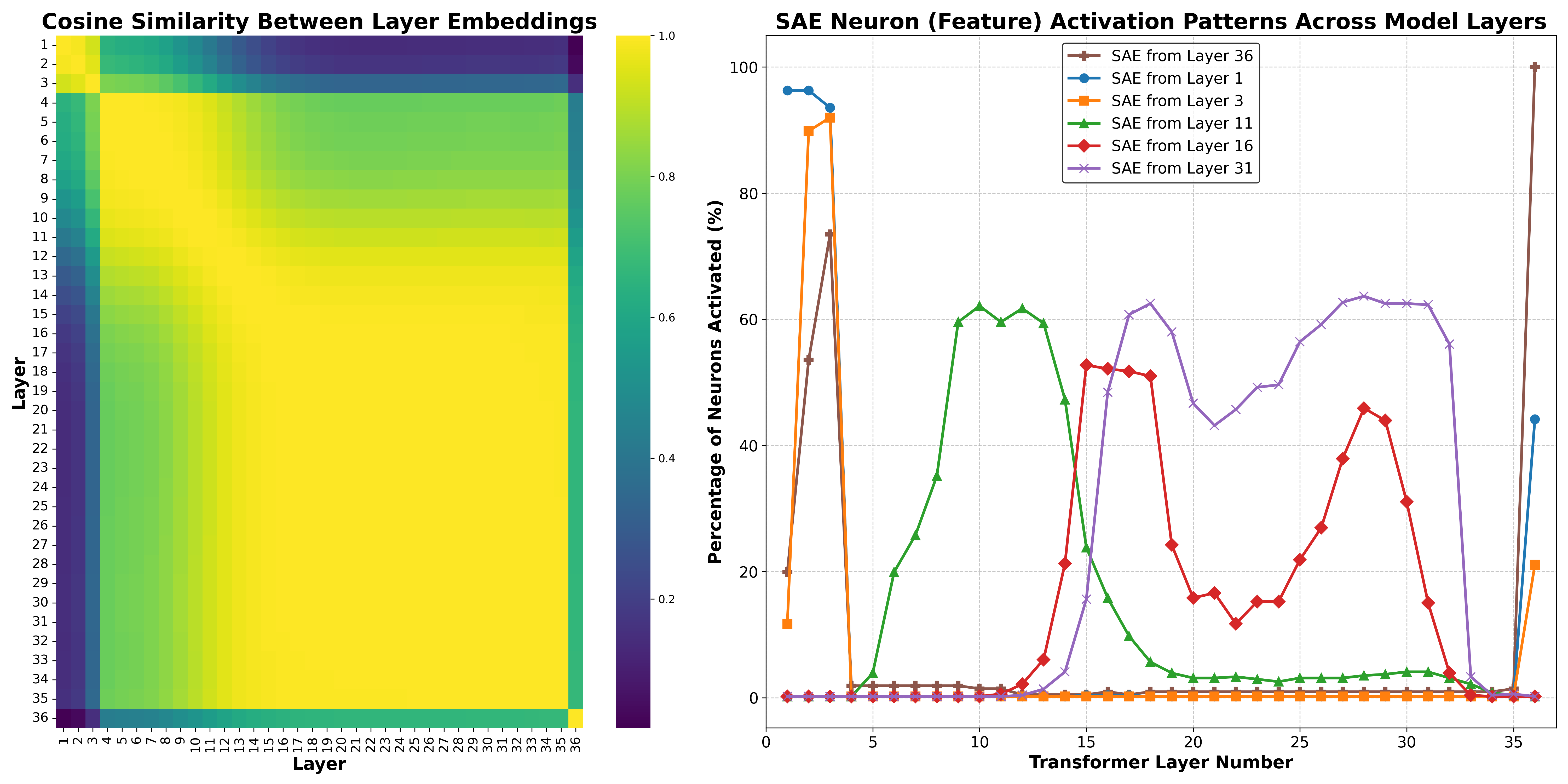}
    \caption{Cosine similarity (left) reveals three macro blocks (embedding, processing core, output), while SAE probes (right) show that feature directions are highly local in depth—with a striking echo between the first and last layers—indicating that the network temporarily hides early lexical axes during computation before restoring them for final prediction.}
    \label{fig:scale_ana_1}
\end{figure}

Figure~\ref{fig:scale_ana_1} illustrates the internal representational dynamics of the model using global cosine similarity between layers' embeddings (left panel) and sparse-autoencoder (SAE) feature-activation probes trained on selected layers (right panel). The cosine similarity heatmap identifies three distinct representational blocks across the model's depth: the input-like block (Layers 1–3), closely aligned with the token-embedding space; the processing core (Layers 4–35), characterized by internal coherence yet significantly different from both input and output spaces; and the output block (Layers 36), realigning representations toward the prediction task.

Complementing this global perspective, SAE probes trained on individual layers offer localized insights by highlighting specific representational directions active at different depths. Notably, early-layer SAEs (Layers 0 and 2) exhibit high initial activations that sharply decline across intermediate layers before partially re-emerging at later layers, particularly at the final output stage. This reactivation underscores a semantic linkage, suggesting that early semantic and lexical features, initially presumed to be transient, actually reappear strategically at later stages. Mid-layer SAEs (Layers 10 and 15) reveal strongly localized activations around their training layers, emphasizing transient, depth-specific representational roles. In contrast, upper-layer SAE (Layer 30) captures broader, sustained activations at higher layers, indicating stable high-level representations crucial for model predictions.

The observed re-emergence of early-layer features in later layers challenges the initial hypothesis of spatially "un-knowledgeable" representations, demonstrating a complex semantic continuity between early and late stages. It can happen for a number of reasons such as early and late layers capturing token specific details, whereas the processing core is distilling and 


For brevity and clarity, detailed per-concept analyses like section \ref{sec:time-concepts} is reported in the Appendix.
\section{Feature Emergence at Scale}
\label{sec:space_ana}

\subsection{Cross–scale alignment}
To embed all checkpoints in a common feature space we apply an \emph{orthogonal Procrustes} transformation \cite{schoenemann1966procrustes, smith2019neuralalignment}.  
For each model \(m\) with hidden-state width \(d_m\) and activation matrix \(X_{m}\in\mathbb{R}^{N\times d_m}\) we solve
\begin{equation}
  W_m^\star = \arg\min_{W\in O(d_m,5120)}
  \bigl\lVert X_{m} W - X_{12\text{B}}\bigr\rVert_F ,
  \label{eq:procrustes}
\end{equation}
where \(X_{12\text{B}}\in\mathbb{R}^{N\times5120}\) is the reference matrix from the 12-Billion-parameter checkpoint.
Zero-padding each \(X_{m}\) to 5120 dimensions would introduce empty coordinates and bias similarity metrics; solving~\eqref{eq:procrustes} keeps the full rank of every smaller model while rotating it into the reference basis.

The optimization is applied to activations of \(N=26{,}074\) evaluation sequences for checkpoints at 14 M, 31 M, 70 M, 160 M, 410 M, 1 B, 1.4 B, 2.8 B, and 6.9 B parameters, whose hidden widths are
\(\{128,\,256,\,512,\,768,\,1024,\,2048,\,2560,\,4096\}\); the 12 B checkpoint supplies the 5120-dimensional reference space.
After alignment each projected matrix \(X_{m}W_m^\star\) shares this basis, allowing direct comparison of feature activations.We assess the fidelity of every projection with two geometry-preservation scores:
(i) \textbf{linear CKA}, which compares dot-product structure, and
(ii) \textbf{pairwise cosine matrix correlation}, which preserves local angular relationships. Both metrics are reported for all scale points in the appendix.  High values in both panels indicate that the Procrustes rotation maintains the global and local geometry of the original activations.


\subsection{Scale–wise Concept Activation}

\begin{figure}[h]
    \centering
    \includegraphics[width=\linewidth]{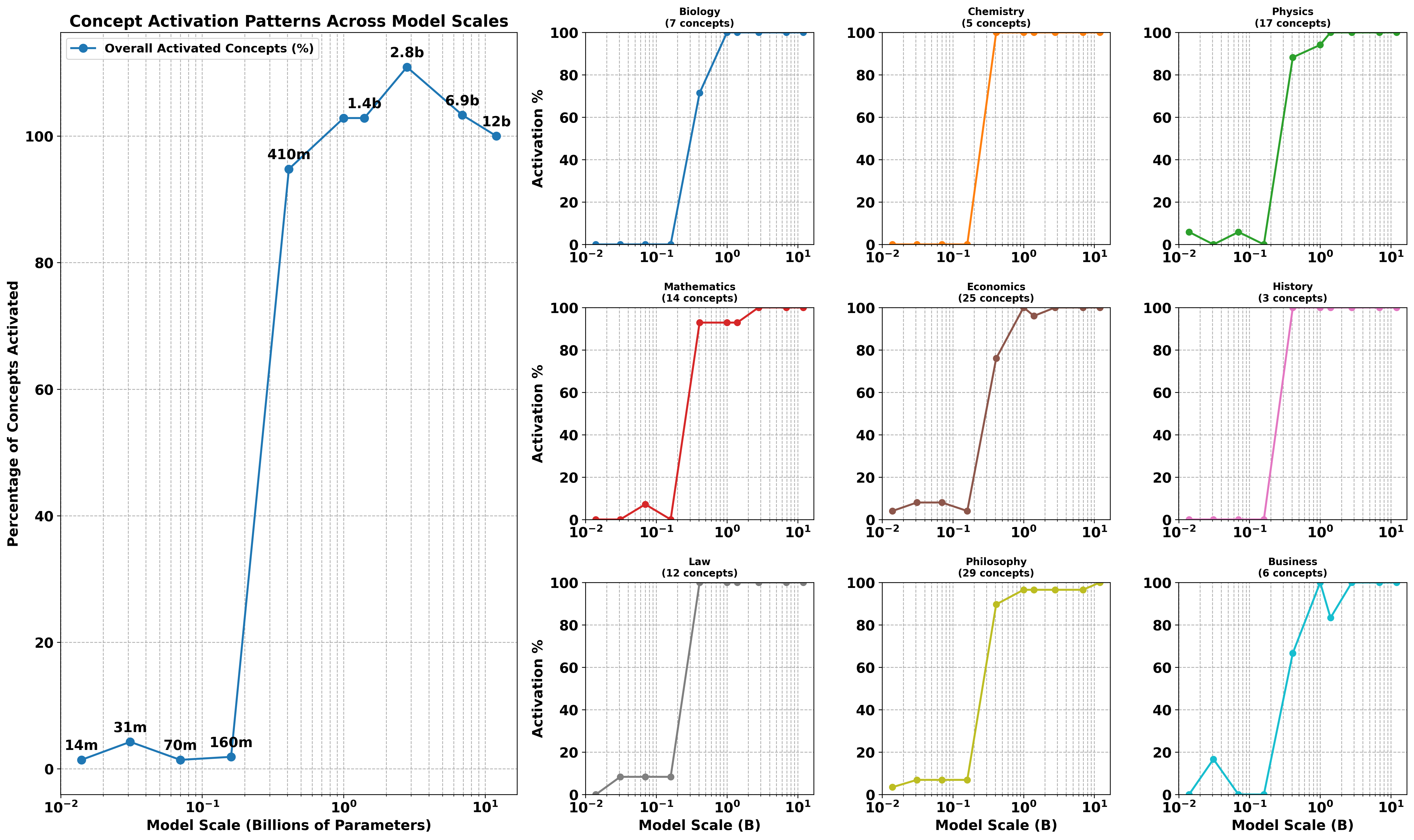}
    \caption{Concept-activation saturation with model scale. 
    \textbf{Left:} percentage of all concepts or features which is activated for each Pythia checkpoint from 14\,M to 12\,B parameters (log scale). 
A single inflection between the 160\,M and 410\,M models raises activation from $<\!5\%$ to $\approx95\%$, after which the curve plateaus. 
\textbf{Right:} per-domain activation profiles show similar critical points for most areas, while Business concepts rise more gradually.}
    \label{fig:scale_ana_2}
\end{figure}

Figure~\ref{fig:scale_ana_2} plots the percentage of concepts activated as a function of model size (log-scaled parameter count, left) together with per–domain traces (right).  
Domain definitions follow the \textsc{EyeSee} taxonomy introduced in Section~\ref{sec:eyesee}.

\paragraph{Global pattern.}
Models below \(200\) M parameters activate fewer than \(5\%\) of the labelled concepts.  
A single transition—between the 160 M and 410 M checkpoints—raises the activation rate by \(+92.9\) pp to \(\sim\!95\%\).  
Beyond this point activation saturates, peaking at 2.8 B parameters and remaining above \(98\%\) for all larger scales considered.

\paragraph{Domain-level thresholds.}
Most subject areas share the same critical jump at 410 M:
\begin{itemize}
    \item \textbf{STEM:} Biology, Chemistry, Physics, and Mathematics all move from \(\le10\%\) at 160 M to \(\ge\!90\%\) activation at 410 M; minor gains follow up to 1 B, after which they saturate.
    \item \textbf{Social sciences:} Economics and Law rise similarly (\(+72\) pp and \(+91.7\) pp respectively) at 410 M, reaching full activation by 1 B.
    \item \textbf{History:} shifts directly from 0 
    \item \textbf{Philosophy:} increases by \(+82.8\) pp at 410 M and stabilises thereafter.
    \item \textbf{Business:} diverges from the pattern: minimal activation already appears at 14–31 M, oscillates at intermediate scales, and only reaches stable activation (\(\ge\!90\%\)) once the model exceeds 1 B parameters.
\end{itemize}

\paragraph{Interpretation.}
The uniform 410 M threshold suggests a capacity requirement for storing the categorical concepts defined in \textsc{EyeSee}.  
Smaller models allocate parameters to high-frequency surface statistics but cannot sustain the richer feature subspaces captured by our activation metric.  
Business concepts appear earlier, possibly due to the higher lexical frequency of business-related terms in the pre-training corpus, but still require larger scales for consistent coverage.

\section{Related Work}
Emergence has mostly been studied along a \emph{single} axis.  
\textbf{Time.}  Checkpoint-level analyses follow when circuits appear or phase-shift (e.g., induction heads \citep{olsson2022incontext}; feature-coherence phases \citep{xu2024tracking}).  
\textbf{Depth.}  Layer probes reveal a lexical→syntactic→semantic hierarchy \citep{jawahar2019berthierarchy} and show concept complexity rising toward upper layers \citep{jin2024conceptdepth}; recent alignment methods match features across neighbouring layers \citep{balagansky2025saematch}.  
\textbf{Scale.}  Parameter-sweep studies document capability jumps at size thresholds \citep{kaplan2020scaling,wei2022emergent} and find that larger models contain more monosemantic features \citep{templeton2024monosemantic}.  

\textbf{Our contribution.}  We jointly track the \emph{same} sparse-auto-encoder features across time, depth \emph{and} scale in a single model family, revealing cross-axis interactions invisible to single-axis work. 

\section{Conclusion}

In this study, we investigated the emergence and evolution of interpretable categorical features within large language models (LLMs) across the complementary axes of time (training checkpoints), space (positions in the transformer stack), and scale (parameter count). Using sparse autoencoders as mechanistic interpretability tools, we demonstrated their effectiveness in identifying semantically meaningful features within model activations. Our fine-grained analysis across domains revealed a structured and progressive activation pattern, where different knowledge areas emerge and stabilize at distinct points during training, varying both temporally and by model scale. Additionally, we observed that features identified in early layers of the model often re-emerge at later stages, challenging the hypothesis of spatially uniform "unknowledgeable" representations.

Despite the detailed observations provided, this work primarily remains descriptive. One limitation is that we did not conduct a finer-grained analysis to elucidate the underlying reasons for these emergent patterns, such as the specific contributions of training data distributions or internal network circuitry to the observed feature activations. Additionally, our choice of sparse autoencoders and specific concept matching criteria impose methodological constraints, potentially missing other important feature dynamics. These aspects represent key avenues for future research.

On a broader, philosophical note, our findings affirm the common hypothesis that knowledge indeed emerges at particular points in time and scale within LLMs. However, the dynamics across the spatial dimension—where semantic features appear transiently, disappear, and then re-emerge—highlight a more nuanced reality. This indicates that while knowledge acquisition aligns well with intuitive expectations temporally and at scale, its spatial organization within neural architectures may not conform to straightforward hypotheses, underscoring the complexity inherent in interpreting neural representation spaces.

\newpage
\bibliographystyle{unsrtnat}
\bibliography{neurips_2025.bbl}

\newpage
\appendix
\section{Appendix}

\subsection{Complete ranked list of different EyeSee concepts }

\begin{center}
\setlength{\LTcapwidth}{\textwidth}
\begin{longtable}{lcccc}
\caption{Complete catalogue of subject-aligned neurons obtained from the $h=512,\,k=1$ SAE.  
Cosine similarity is computed between the MPNet embedding of the AutoInterp label and the subject query;  
F1 is the verification fidelity.  Data source: AutoInterp output.} \label{tab:all-concept-neurons} \\[0.4em]
\toprule
\textbf{Subject} & \textbf{Neuron ID} & \textbf{AutoInterp Label} & \textbf{Cos.\ Sim.} & \textbf{F1} \\
\midrule
\endfirsthead

\multicolumn{5}{c}%
{{\tablename\ \thetable{} -- continued from previous page}}\\[0.2em]
\toprule
\textbf{Subject} & \textbf{Neuron ID} & \textbf{AutoInterp Label} & \textbf{Cos.\ Sim.} & \textbf{F1} \\
\midrule
\endhead

\midrule \multicolumn{5}{r}{{Continued on next page}}\\
\endfoot

\bottomrule
\endlastfoot
Biology & 38  & Natural Phenomena and Processes       & 0.377 & 0.91 \\
        & 295 & Genetic Variation Source              & 0.366 & 1.00 \\
        & 269 & Human-related Processes               & 0.347 & 0.91 \\
        & 511 & Sexual Dimorphism Selection           & 0.342 & 1.00 \\
        & 386 & Life and Development Concepts         & 0.339 & 1.00 \\
        & 47  & Genetic and Sensory Differences       & 0.331 & 1.00 \\
        & 404 & Fatty Acid Transport                  & 0.302 & 1.00 \\[0.3em]
\midrule
Chemistry & 194 & Paper-related Chemistry             & 0.491 & 1.00 \\
          & 125 & Iron and Silver Chemistry           & 0.406 & 1.00 \\
          & 38  & Natural Phenomena and Processes     & 0.385 & 0.91 \\
          & 483 & Water-related Thermodynamics        & 0.335 & 1.00 \\
          & 126 & Free Radicals and Psychoanalysis    & 0.302 & 1.00 \\[0.3em]
\midrule
Physics & 78  & Newton’s Laws Applications            & 0.513 & 1.00 \\
        & 195 & Mathematical Problem Solving          & 0.425 & 1.00 \\
        & 12  & Real-world Mathematical Applications  & 0.401 & 1.00 \\
        & 403 & Light-related Phenomena               & 0.384 & 1.00 \\
        & 484 & Numerical Problem Solving             & 0.379 & 0.91 \\
        & 38  & Natural Phenomena and Processes       & 0.368 & 0.91 \\
        & 341 & Airflow and Heat Transfer             & 0.363 & 1.00 \\
        & 400 & Logical and Mathematical Concepts     & 0.346 & 0.91 \\
        & 421 & Human Perception and Interaction      & 0.336 & 1.00 \\
        & 183 & Expressing Quantities Mathematically  & 0.323 & 0.91 \\
        & 291 & Mechanical Design Calculations        & 0.318 & 1.00 \\
        & 135 & Passive Transport                     & 0.317 & 1.00 \\
        & 282 & Heat Transfer and Efficiency          & 0.312 & 1.00 \\
        & 324 & Numerical Computation Problems        & 0.312 & 1.00 \\
        & 483 & Water-related Thermodynamics          & 0.309 & 1.00 \\
        & 473 & Electron and Electromagnetic Concepts & 0.306 & 1.00 \\
        & 392 & Performance and Analysis              & 0.302 & 1.00 \\[0.3em]
\midrule
Mathematics & 195 & Mathematical Problem Solving         & 0.707 & 1.00 \\
            & 400 & Logical and Mathematical Concepts    & 0.590 & 0.91 \\
            & 12  & Real-world Mathematical Applications & 0.549 & 1.00 \\
            & 130 & Simple Arithmetic Problems           & 0.528 & 1.00 \\
            & 183 & Expressing Quantities Mathematically & 0.492 & 0.91 \\
            & 484 & Numerical Problem Solving            & 0.450 & 0.91 \\
            & 324 & Numerical Computation Problems       & 0.377 & 1.00 \\
            & 58  & Comparative Analysis                 & 0.324 & 0.91 \\
            & 392 & Performance and Analysis             & 0.319 & 1.00 \\
            & 38  & Natural Phenomena and Processes      & 0.316 & 0.91 \\
            & 234 & Conditional Reasoning                & 0.311 & 0.91 \\
            & 421 & Human Perception and Interaction     & 0.306 & 1.00 \\
            & 377 & Complex Procedural Knowledge         & 0.304 & 1.00 \\
            & 276 & Logical Reasoning in Statements      & 0.301 & 0.91 \\[0.3em]
\midrule
Economics & 367 & Business and Economic Dynamics        & 0.600 & 1.00 \\
          & 55  & Human Behavior and Decision-Making    & 0.485 & 0.91 \\
          & 218 & Economic Growth Factors               & 0.480 & 1.00 \\
          & 4   & Social Dynamics and Influence         & 0.476 & 1.00 \\
          & 305 & Decision-Making in Institutions       & 0.418 & 1.00 \\
          & 23  & Environmental Ethics                  & 0.405 & 1.00 \\
          & 269 & Human-related Processes               & 0.371 & 0.91 \\
          & 457 & Mill's Utilitarian Philosophy         & 0.369 & 1.00 \\
          & 195 & Mathematical Problem Solving          & 0.368 & 1.00 \\
          & 492 & Interest Group Influence              & 0.357 & 1.00 \\
          & 294 & Long-term Consequences                & 0.349 & 1.00 \\
          & 264 & Cost and Tax Analysis                 & 0.348 & 1.00 \\
          & 456 & Distribution Channels                 & 0.346 & 1.00 \\
          & 446 & Ethical and Cultural Positions        & 0.344 & 1.00 \\
          & 488 & Psychological Concepts and Ethics     & 0.342 & 1.00 \\
          & 58  & Comparative Analysis                  & 0.340 & 0.91 \\
          & 400 & Logical and Mathematical Concepts     & 0.322 & 0.91 \\
          & 354 & Contrast and Comparison               & 0.319 & 0.91 \\
          & 386 & Life and Development Concepts         & 0.318 & 1.00 \\
          & 350 & Ethical and Philosophical Concepts    & 0.315 & 0.91 \\
          & 370 & Family-related Decision Making        & 0.312 & 1.00 \\
          & 245 & Critique of Consequentialism          & 0.309 & 1.00 \\
          & 61  & Decision-Making Scenarios             & 0.306 & 0.91 \\
          & 421 & Human Perception and Interaction      & 0.303 & 1.00 \\
          & 298 & “Comparative Analysis Questions”      & 0.302 & 1.00 \\[0.3em]
\midrule
History & 4   & Social Dynamics and Influence           & 0.387 & 1.00 \\
        & 58  & Comparative Analysis                    & 0.322 & 0.91 \\
        & 446 & Ethical and Cultural Positions          & 0.320 & 1.00 \\[0.3em]
\midrule
Law & 510 & Legal Decision-Making Criteria            & 0.454 & 1.00 \\
    & 57  & Warranty Types in Law                     & 0.377 & 1.00 \\
    & 305 & Decision-Making in Institutions           & 0.345 & 1.00 \\
    & 50  & Ethical and Moral Concepts                & 0.341 & 1.00 \\
    & 446 & Ethical and Cultural Positions            & 0.337 & 1.00 \\
    & 4   & Social Dynamics and Influence             & 0.324 & 1.00 \\
    & 105 & Land and Property Rights                  & 0.324 & 1.00 \\
    & 440 & Warrantless Searches and Privacy          & 0.323 & 1.00 \\
    & 23  & Environmental Ethics                      & 0.321 & 1.00 \\
    & 187 & Virtue Ethics and Morality                & 0.314 & 1.00 \\
    & 400 & Logical and Mathematical Concepts         & 0.311 & 0.91 \\
    & 425 & Congressional Powers and Limitations      & 0.304 & 1.00 \\[0.3em]
\midrule
Philosophy & 350 & Ethical and Philosophical Concepts      & 0.606 & 0.91 \\
           & 285 & Socratic Philosophy Concepts            & 0.477 & 1.00 \\
           & 446 & Ethical and Cultural Positions          & 0.462 & 1.00 \\
           & 187 & Virtue Ethics and Morality              & 0.453 & 1.00 \\
           & 23  & Environmental Ethics                    & 0.453 & 1.00 \\
           & 50  & Ethical and Moral Concepts              & 0.452 & 1.00 \\
           & 488 & Psychological Concepts and Ethics       & 0.428 & 1.00 \\
           & 496 & Kantian Ethics Principles               & 0.397 & 1.00 \\
           & 400 & Logical and Mathematical Concepts       & 0.391 & 0.91 \\
           & 4   & Social Dynamics and Influence           & 0.387 & 1.00 \\
           & 55  & Human Behavior and Decision-Making      & 0.387 & 0.91 \\
           & 305 & Decision-Making in Institutions         & 0.384 & 1.00 \\
           & 69  & Ethical Debates on Euthanasia           & 0.379 & 1.00 \\
           & 245 & Critique of Consequentialism            & 0.375 & 1.00 \\
           & 457 & Mill's Utilitarian Philosophy           & 0.361 & 1.00 \\
           & 386 & Life and Development Concepts           & 0.359 & 1.00 \\
           & 18  & Moral Complexity in Abortion            & 0.357 & 1.00 \\
           & 269 & Human-related Processes                 & 0.342 & 0.91 \\
           & 360 & Evidence-Based Reasoning                & 0.340 & 0.91 \\
           & 421 & Human Perception and Interaction        & 0.333 & 1.00 \\
           & 58  & Comparative Analysis                    & 0.327 & 0.91 \\
           & 306 & Self-related Psychological Concepts     & 0.326 & 1.00 \\
           & 195 & Mathematical Problem Solving            & 0.323 & 1.00 \\
           & 276 & Logical Reasoning in Statements         & 0.322 & 0.91 \\
           & 298 & “Comparative Analysis Questions”        & 0.321 & 1.00 \\
           & 157 & Ethical Dilemmas in Abortion            & 0.319 & 1.00 \\
           & 354 & Contrast and Comparison                 & 0.313 & 0.91 \\
           & 98  & Carl Jung Concepts                      & 0.305 & 1.00 \\
           & 104 & Ethical Dilemmas in Therapy             & 0.303 & 1.00 \\[0.3em]
\midrule
Business & 367 & Business and Economic Dynamics         & 0.548 & 1.00 \\
         & 456 & Distribution Channels                  & 0.419 & 1.00 \\
         & 55  & Human Behavior and Decision-Making      & 0.314 & 0.91 \\
         & 269 & Human-related Processes                 & 0.312 & 0.91 \\
         & 492 & Interest Group Influence                & 0.310 & 1.00 \\
         & 4   & Social Dynamics and Influence           & 0.308 & 1.00 \\
\end{longtable}
\end{center}

\newpage
\subsection{\textsc{EyeSee} analysis for space}
\label{sec:space_ana}

Across depth the subject‐aligned neurons exhibit the same “disappear-and-return” motif seen in our global SAE study: most topics fire strongly in the first three blocks, fall almost completely silent throughout the mid-stack, and then re-emerge when the Layer-36 SAE is probed. However, the precise silence window and reactivation point vary by domain—e.g., History and Business neurons vanish after Block 3 but resurface sharply at the final block, whereas Philosophy keeps a faint 3–4 \% signal until about Block 11 before dropping off and later returning. These differences suggest that each subject’s semantic cues are distilled and re-inserted on slightly different schedules, yet the overarching early-hide-late-recall pattern remains consistent.

\begin{figure}[h]
    \centering
    \includegraphics[width=\linewidth]{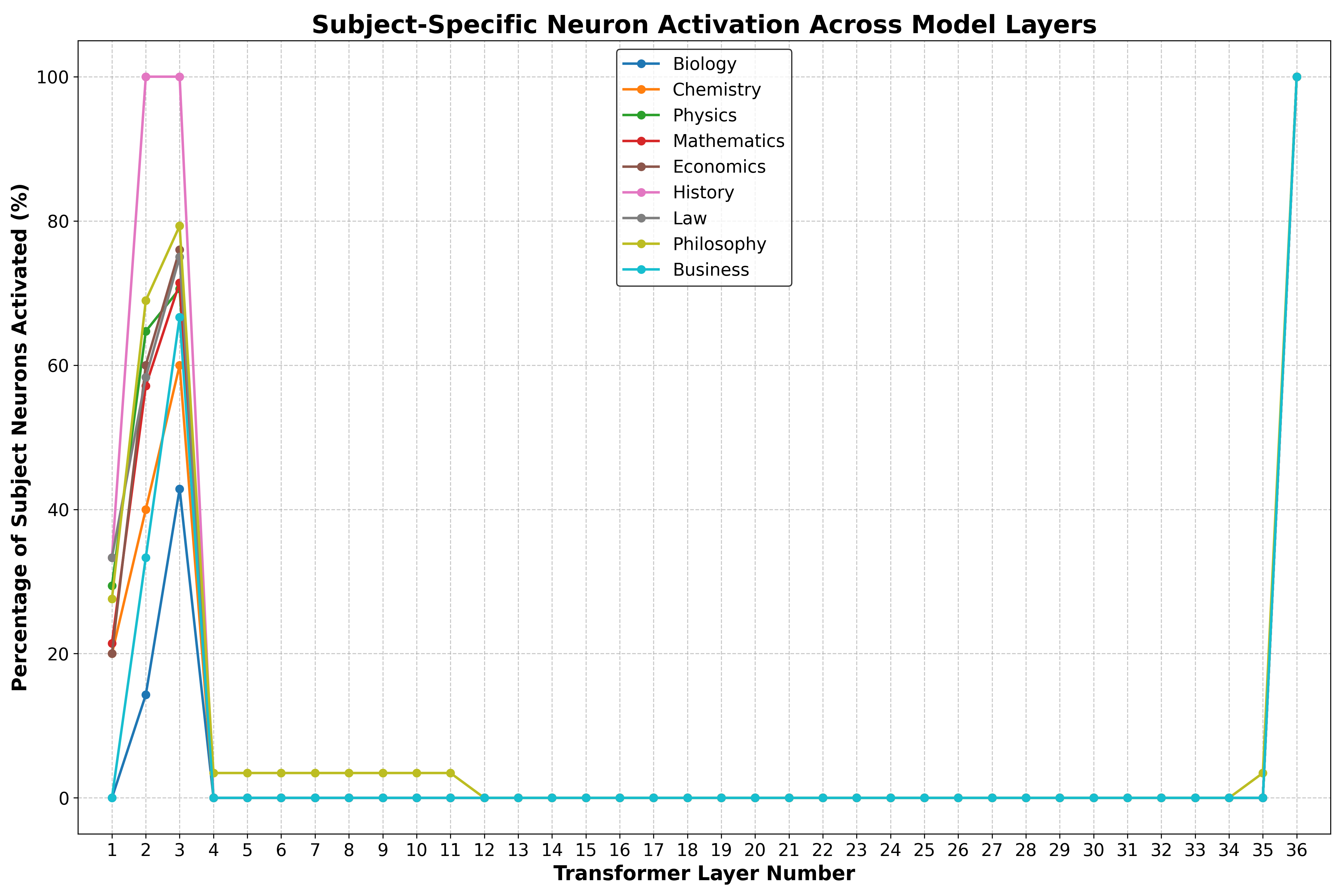}
    \caption{Percentage of high-fidelity neurons (F1 $\ge 0.9$) that fire for each subject in every transformer block of the 12-B Pythia model.  Values are averaged over the combined \textsc{MMLU} + \textsc{MMLU-Pro} prompt set.}
    \label{fig:subject_layer_curve}
\end{figure}
\newpage

\subsection{Performance of Procrustes Rotation Alignment}

\begin{figure}[h]
    \centering
    \includegraphics[width=\linewidth]{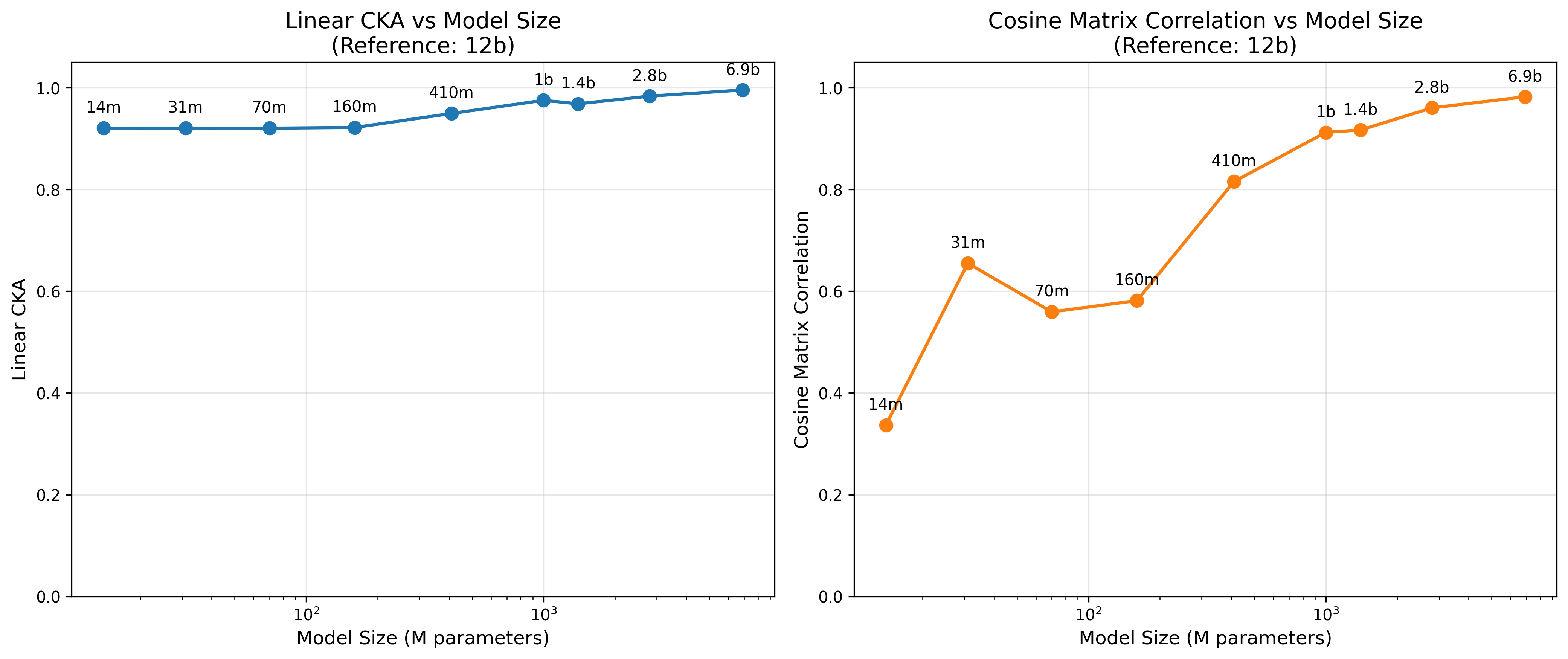}
    \caption{Alignment quality across model scale.
           Left: linear CKA between projected activations \(X_{m}W_m^\star\) and the reference \(X_{12\text{B}}\).
           Right: pairwise cosine matrix correlation for the same pairs.}
    \label{fig:scale_alignment}
\end{figure}

Figure~\ref{fig:scale_alignment} quantifies how closely each checkpoint matches the 12-B reference after Procrustes rotation.  
The left panel shows that \emph{linear CKA} stays above \(0.90\) even for the smallest 14-M model and increases monotonically to \(\approx 0.99\) at 6.9 B, indicating strong preservation of the \emph{global} dot-product structure across scale.  
The right panel plots the \emph{pairwise cosine matrix correlation}, which starts lower (0.34 for 14 M) because local neighbourhood geometry differs more in tiny models, but rises sharply from the 410-M checkpoint onward and exceeds \(0.95\) for all models with \(\ge 2.8\) B parameters.  
Together, the two metrics confirm that the orthogonal alignment retains both coarse and fine-grained geometry, with local agreement improving consistently as model size grows.

\label{appen:PRA_perf}


\end{document}